\documentclass[a4paper,twoside]{article}

\usepackage{epsfig}
\usepackage{subcaption}
\usepackage{calc}
\usepackage{amssymb}
\usepackage{amstext}
\usepackage{amsmath}
\usepackage{amsthm}
\usepackage{multicol}
\usepackage{pslatex}
\usepackage{apalike}
\usepackage{algorithm2e}
\usepackage[bottom]{footmisc}
\usepackage{url}
\usepackage{xcolor}
\usepackage{todonotes}
\usepackage{siunitx}
\usepackage{subcaption}
\usepackage{multirow}
\usepackage{pgfplots}
\pgfplotsset{compat=newest}
\usepackage[breaklinks=true]{hyperref}
\usepackage{tikz}
\usetikzlibrary{shapes.geometric, arrows}
\usetikzlibrary{external}
\usepackage{SCITEPRESS}     % Please add other packages that you may need BEFORE the SCITEPRESS.sty package.

\definecolor{TUMBlue}{rgb}{0.0, 0.40, 0.74}
\definecolor{TUMGray1}{rgb}{0.2, 0.2, 0.2}
\definecolor{TUMGray3}{rgb}{0.8, 0.8, 0.8}
\definecolor{TUMBlue4}{rgb}{0.60, 0.78, 0.92}
\definecolor{TUMIvory}{rgb}{0.85, 0.84, 0.80}
\definecolor{TUMOrange}{rgb}{0.89, 0.45, 0.13}
\definecolor{TUMGreen}{rgb}{0.64, 0.68, 0.0}
\definecolor{TUMGreenWeb}{rgb}{0.62, 0.73, 0.21}
\definecolor{TUMRedWeb}{rgb}{0.92, 0.45, 0.22}

\begin{document}

\title{FlexCloud: Direct, Modular Georeferencing and Drift-Correction of Point Cloud Maps}

\author{\authorname{Maximilian Leitenstern\sup{1}\orcidAuthor{0009-0008-6436-7967}, Marko Alten\sup{1}, Christian Bolea-Schaser\sup{1}, Dominik Kulmer\sup{1}\orcidAuthor{0000-0001-7886-7550}, Marcel Weinmann\sup{1}\orcidAuthor{0009-0008-7174-4732}, and Markus Lienkamp\sup{1}\orcidAuthor{0000-0002-9263-5323}}
\affiliation{\sup{1}Institute of Automotive Technology, Technical University of Munich, Boltzmannstraße 15, 85748 Garching, Germany}
\email{maxi.leitenstern@tum.de}}
\abstract{Current software stacks for real-world applications of autonomous driving leverage map information to ensure reliable localization, path planning, and motion prediction. An important field of research is the generation of point cloud maps, referring to the topic of simultaneous localization and mapping~(SLAM). As most recent developments do not include global position data, the resulting point cloud maps suffer from internal distortion and missing georeferencing, preventing their use for map-based localization approaches. Therefore, we propose \textit{FlexCloud} for an automatic georeferencing of point cloud maps created from SLAM. Our approach is designed to work modularly with different SLAM methods, utilizing only the generated local point cloud map and its odometry. Using the corresponding GNSS positions enables direct georeferencing without additional control points. By leveraging a 3D rubber-sheet transformation, we can correct distortions within the map caused by long-term drift while maintaining its structure. Our approach enables the creation of consistent, globally referenced point cloud maps from data collected by a mobile mapping system~(MMS). The source code of our work is available at \url{https://github.com/TUMFTM/FlexCloud}.}

\onecolumn \maketitle \normalsize \setcounter{footnote}{0} \vfill
%
% ======================= %
% === Introduction === %
% ======================= %
\section{\uppercase{Introduction}}
\label{sec:introduction}
So-called High-Definition (HD) maps are considered a key technology to enable real-world autonomous driving~\cite{seif_autonomous_2016} as they provide essential prior information on static surroundings of the vehicle, including temporally occluded objects~\cite{bao_high-definition_2022,srinara_strategy_2023}. Therefore, HD maps can be seen as an additional sensor facilitating localization and motion planning by combination with other sensors, such as a Global Navigation Satellite System (GNSS), an Inertial Measurement Unit (IMU), or a Light Detection and Ranging (LiDAR) system. Unlike traditional Standard-Definition (SD) maps used for navigation, HD maps feature an accuracy of \SIrange[]{10}{20}{\centi \meter}~\cite{jeong_tutorial_2022}. Depending on the application's requirements, different map representations evolved~\cite{khoche_semantic_2022}. Hence, an HD map is usually split into a point cloud map (PCM) and a vector map~\cite{jeong_tutorial_2022}.
\tikzstyle{process} = [rectangle, rounded corners, text width=2.2cm, minimum height=0.5cm, text centered, draw=black, fill=TUMBlue!20, font=\small]
\tikzstyle{arrow} = [thick,->,>=stealth]
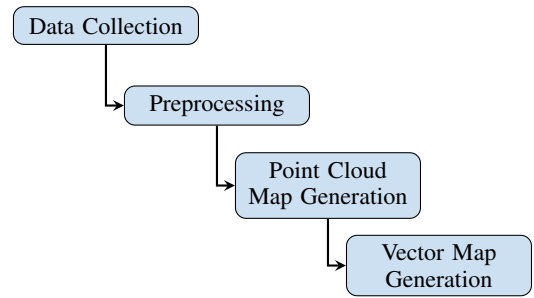
\begin{figure}[!htb]
    \centering
    \begin{tikzpicture}[node distance=1.5cm]
    
        % Nodes
        \node (data_collection) [process] {Data Collection};
        \node (data_preprocessing) [process, below right of=data_collection, xshift=0.4cm] {Preprocessing};
        \node (point_cloud) [process, below right of=data_preprocessing, xshift=0.4cm] {Point Cloud Map Generation};
        \node (vector_map) [process, below right of=point_cloud, xshift=0.4cm] {Vector Map Generation};
        
        % Arrows
        \draw [arrow] (data_collection) |- (data_preprocessing);
        \draw [arrow] (data_preprocessing) |- (point_cloud);
        \draw [arrow] (point_cloud) |- (vector_map);
    
    \end{tikzpicture}
    \caption{Steps for HD map generation (extended from~\cite{srinara_strategy_2023}).}
    \label{fig:hdmapgeneration}
\end{figure}
While a PCM depicts the 3D structure of objects by individual points, a vector map represents the location of relevant objects (e.g., lanes, traffic signs) by combining points, lines, and polygons~\cite{jeong_tutorial_2022}. Leveraging the strengths of the respective map format, the PCM is used for localization while the vector map assists during motion planning~\cite{jeong_tutorial_2022}. \\%
HD map generation can be classified into offline and online approaches~\cite{bao_high-definition_2022}. Following the architecture of the presented approach, this work classifies as a tool to enhance the offline generation of HD maps. This process generally comprises the steps shown in~\autoref{fig:hdmapgeneration}. It starts with data collection using a MMS~\cite{bao_review_2023}. During data preprocessing, several algorithms are applied to the collected LiDAR data to facilitate the creation of the PCM in the following step~\cite{bao_review_2023,srinara_strategy_2023}. After that, a PCM is created by aligning the individual LiDAR scans, usually using scan registration techniques. In the last step, the vector map is created from the PCM, completing the HD map. \\%
The generation of PCMs from sensor data is extensively studied in the literature, usually known as SLAM. However, recent developments within this field focus on approaches that only use LiDAR and, occasionally, IMU data. Although this reduces the required data sources and avoids calibration or time synchronization issues, the created local PCM inherently suffers from long-term drift and missing global reference. Hence, georeferencing and correction for the long-term drift are necessary to leverage these maps for localization, especially if global GNSS coordinates are used with a LiDAR-based approach. 
In a previous work of ours~\cite{leitenstern_flexmap_2024}, we presented an approach to overcome this issue. However, our approach still suffered from the required manual effort and the limitation to 2D. Therefore, in this paper, we present \textit{FlexCloud} as a modular pipeline to overcome these limitations. Our contributions are as follows:
\begin{itemize}
    \item We propose a novel, modular approach to automatically georeference local PCMs created from SLAM.
    \item We use an interpolation-based strategy to automatically determine control points (CPs) from the corresponding GNSS positions, considering their uncertainty.
    \item We implement a 3D rubber-sheet transformation with Delaunay triangulation to effectively correct the long-term drift of the PCM while maintaining its structure and continuity.
    \item We open-source our pipeline as a standalone \textit{ROS 2}-package upon acceptance of our work.
\end{itemize}
We evaluate the results of our pipeline on self-recorded data at the Yas Marina Circuit (YMC) in Abu Dhabi, UAE. Additionally, we present the results for sequence 00 of the KITTI vision benchmark suite~\cite{geiger_are_2012,geiger_vision_2013} to demonstrate its generalizability.
%
% ======================= %
% === Related Work    === %
% ======================= %
\section{\uppercase{Related Work}}%
\label{sec:related}%
This chapter classifies different methods for georeferencing of mapping data and presents related work regarding the creation of PCMs.
\subsection{Georeferencing of Mapping Data}
\label{subsec:georeferencing}
Generally, georeferencing describes the problem of finding the transformation parameters of a local set of points to a common global reference system, such as WGS84~\cite{paffenholz_direct_2012,lasse_klingbeil_georeferencing_2023}. Initially, a LiDAR captures point clouds in a local coordinate system. With scan registration techniques, the relative transformation between single scans can be computed, and thus, the scans can be assembled to a local PCM. The transformation of this PCM to a global frame is referred to as georeferencing and is described by the following equation~\cite{wilkinson_novel_2010}:
\begin{equation}
    x^{G} = s\,R_{S}^{G}\,x^{S} + t_{S}^{G}.
    \label{eq:georef_def}
\end{equation}
The transformation of a point $x^{S}$ in the PCM consists of the rotation $R_{S}^{G}$, the translation $t_{S}^{G}$ and the scaling parameter $s$, with the indices $S$ and $G$ referring to the local, LiDAR frame and the global frame, respectively. According to Paffenholz~\cite{paffenholz_direct_2012}, georeferencing techniques can be classified as follows:
\begin{itemize}
    \item \textbf{Indirect Georeferencing:} \\%
    These methods use well-spread ground CP, whose coordinates are known locally and globally. The coordinates may be obtained by detection approaches and surveys in local and global frames. The additional effort needed to capture the ground CP is a drawback, making the method time-consuming.
    \item \textbf{Direct Georeferencing:} \\%
    This approach utilizes data from external sensors, usually a GNSS receiver, attached to the LiDAR during data acquisition. Thus, the transformation parameters are directly available from the data acquisition.
    \item \textbf{Data-driven Georeferencing:} \\%
    These methods leverage existing georeferenced data to match the newly acquired data. However, this requires a preceding indirect or direct georeferencing to generate this data.
\end{itemize}
As our approach classifies as a direct method for georeferencing, the following section reviews existing approaches within this field.
\subsection{Direct Point Cloud Georeferencing}
\label{subsec:pointcloudmapgeneration}
In 2005, Schuhmacher and Böhm~\cite{schuhmacher_georeferencing_2005} compared different techniques for georeferencing in architectural modeling applications. For direct georeferencing, they combined a low-cost GNSS system with a digital compass to get the translation and rotation parameters of the transformation, respectively. Wilkinson et al.~\cite{wilkinson_novel_2010} utilize a dual-antenna GNSS system to capture the orientation of the LiDAR. As the accuracy suffers in bad GNSS conditions, they combine it with photogrammetric observations of a camera in a subsequent work~\cite{wilkinson_dual-antenna_2015}. Osada et al.~\cite{osada_direct_2017} propose a method for direct georeferencing that leverages an external, global earth gravity model and thus only requires a minimum number of GNSS measurements. They use a least squares method to estimate the transformation parameters from a GNSS position and the vertical deflection components of the LiDARs rotational axis, determined within the gravitational force field of the earth gravity model. \\%
However, the approaches presented so far have in common that they do not consider the mapping system in a mobile operating mode, thus neglecting uncertainties from kinematic measurement noise and geometric falsification offsets~\cite{liu_georeferencing_2021}. To georeference LiDAR data from an MMS, Oria-Aguilera et al.~\cite{oria-aguilera_mobile_2018} leverage a Real-Time Kinematic (RTK) GNSS system, providing centimeter-level accuracy. They use interpolation to relate the LiDAR scans with GNSS poses to account for issues from different frequencies and time synchronization. The GNSS receiver's Inertial Navigation System (INS) is used to directly determine the transformation parameters for georeferencing without using scan registration techniques. A drawback of this method is the reliance on the INS solution and, thus, good GNSS coverage. To this end, Hariz et al.~\cite{hariz_direct_2021} present an approach for creating georeferenced PCMs using SLAM. They use the Cartographer SLAM~\cite{hess_2016}, taking LiDAR point clouds, IMU measurements, and RTK-corrected GNSS positions as input to create submaps, which are later concatenated to a global map. Koide et al.~\cite{koide_portable_2019} present a similar approach, including GNSS constraints in a graph-based SLAM algorithm to generate georeferenced PCMs. \\%
However, many advancements in LiDAR odometry and SLAM concentrate on algorithms that rely only on LiDAR and potentially IMU data, e.g. \textit{LOAM}~\cite{zhang_loam_2014}, \textit{LeGO-LOAM}~\cite{shan_lego-loam_2018}, \textit{SuMa++}~\cite{chen_suma_2019}, \textit{LIO-SAM}~\cite{shan_lio-sam_2020}, \textit{FAST-LIO}~\cite{xu_fast-lio_2021,xu_fast-lio2_2021}, and \textit{KISS-ICP}~\cite{vizzo_kiss-icp_2023}.
As the PCMs generated with these approaches are in a local frame and distorted due to long-term drift, we presented \textit{FlexMap Fusion} in a previous work~\cite{leitenstern_flexmap_2024}. It leverages rubber sheeting, an approach for the alignment of two spatial datasets~\cite{griffin}, based on RTK-corrected GNSS positions and the SLAM trajectory. Although this work demonstrates the general concept of the approach, it is limited to 2D and requires manual effort to select CPs on the GNSS and SLAM trajectories. 
%
% ======================= %
% === Methodology === %
% ======================= %
\section{\uppercase{Methodology}}
\label{sec:meth}%
\begin{figure*}
    \centering
    \includegraphics[width=\textwidth]{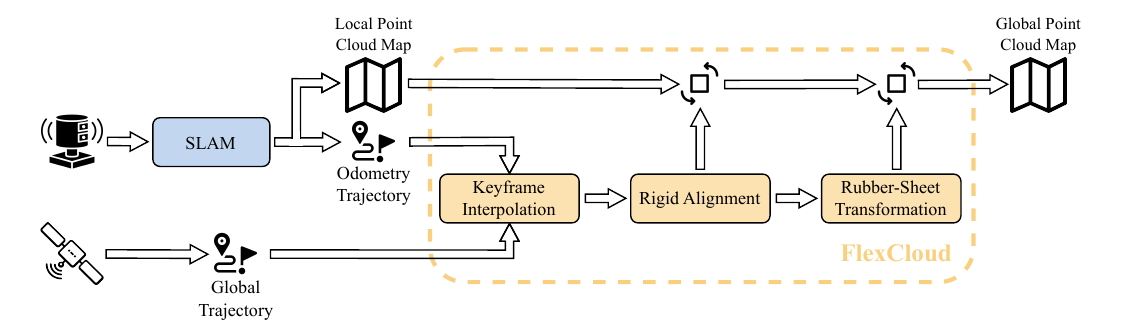}
    \caption{Flowchart illustrating the creation of a global, georeferenced PCM using \textit{FlexCloud}. The computations within the single modules are conducted with the global GNSS trajectory and the local odometry trajectory. The resulting transformations from the \textit{Rigid Alignment} and the \textit{Rubber-Sheet Transformation} are then applied to the PCM.}
    \label{fig:flowchart}
\end{figure*}

To enable georeferencing of PCMs generated from SLAM algorithms, we present \textit{FlexCloud}. It is designed to work with various SLAM approaches and can transform the generated PCM to a global frame while correcting the long-term drift resulting from scan registration errors. \textit{FlexCloud} is intended for use with an MMS, providing LiDAR and GNSS data, preferably from an RTK-corrected GNSS receiver. \autoref{fig:flowchart} illustrates the usage of \textit{FlexCloud} based on these inputs. 
After data recording, an odometry trajectory and a corresponding initial PCM must be generated using SLAM. The experiments in this work are conducted with trajectories generated from \textit{KISS-ICP}~\cite{vizzo_kiss-icp_2023} with additional loop closure constraints inserted using \textit{Interactive SLAM}~\cite{koide_interactive_2021}. This process proved robust and accurate results for PCM generation, as shown in Sauerbeck et al.~\cite{sauerbeck2023}. The resulting odometry trajectory with the corresponding PCM and the GNSS trajectory are the inputs to \textit{FlexCloud}, consisting of the modules \textit{Keyframe Interpolation}, \textit{Rigid Alignment}, and \textit{Rubber-Sheet Transformation}. The following sections describe the modules in detail.
\subsection{\textit{Keyframe Interpolation}}%
\label{subsec:keyframeinterpolation}%
This module determines temporal correspondences between the GNSS and odometry trajectories, later used to select CPs for georeferencing. As a first step, the global GNSS positions are transformed into metric coordinates to facilitate mathematical operations and computations. \textit{FlexCloud} implements a transformation to the East-North-Up (ENU) frame using \textit{GeographicLib}\footnote{https://geographiclib.sourceforge.io/}. However, other transformations, such as the Universal Transverse Mercator (UTM) projection, may be used as well~\cite{groves_principles_2013}.

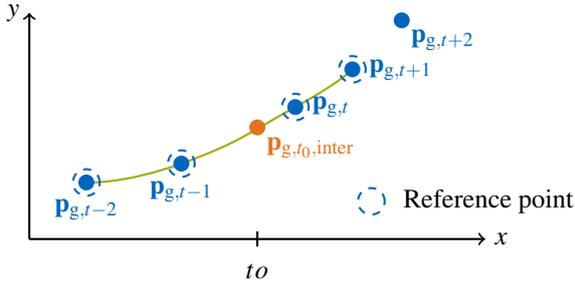
\begin{figure}[!tb]
    \centering
    \begin{tikzpicture}
        % Axis
        \draw[->, thick] (0, 0) -- (6, 0) node[right] {$x$};
        \draw[->, thick] (0, 0) -- (0, 3) node[left] {$y$};

        % Interpolated curve
        \draw[thick, TUMGreen] 
    plot [smooth, tension=1.0] coordinates {(0.75,0.75) (2.0,1.0) (3.5,1.75) (4.25,2.25)};

        % Global points
        \fill[TUMBlue] (0.75, 0.75) circle (3pt) node[below, yshift=-1mm] {$\mathbf{p}_{\mathrm{g},t-2}$};
        \fill[TUMBlue] (2.0, 1.0) circle (3pt) node[below, yshift=-1mm] {$\mathbf{p}_{\mathrm{g},t-1}$};
        \fill[TUMBlue] (3.5, 1.75) circle (3pt) node[right, xshift=+1mm] {$\mathbf{p}_{\mathrm{g},t}$};
        \fill[TUMBlue] (4.25, 2.25) circle (3pt) node[right, xshift=+1mm] {$\mathbf{p}_{\mathrm{g},t+1}$};
        \fill[TUMBlue] (4.9, 2.9) circle (3pt) node[below right] {$\mathbf{p}_{\mathrm{g},t+2}$};

        \fill[TUMOrange] (3, 1.48) circle (3pt) node[below right] {$\mathbf{p}_{\mathrm{g},t_0,\mathrm{inter}}$};

        % Highlighted points with dashed border
        \draw[dashed, thick, TUMBlue] (0.75, 0.75) circle (5pt);
        \draw[dashed, thick, TUMBlue] (2.0, 1.0) circle (5pt);
        \draw[dashed, thick, TUMBlue] (3.5, 1.75) circle (5pt);
        \draw[dashed, thick, TUMBlue] (4.25, 2.25) circle (5pt);

        % Legend
        \begin{scope}[shift={(4.5, 0.5)}] % Adjust the position of the legend
            \draw[dashed, thick, TUMBlue] (0, 0) circle (5pt);
            \node[right] at (0.3, 0) {Reference point};
        \end{scope}

        % X-axis tick with text box
        \draw[thick] (3, 0.1) -- (3, -0.1); % Tick mark
        \node[below] at (3, -0.2) {$to$}; % Text box below the tick
    
    \end{tikzpicture}
    \caption{Principle of \textit{Keyframe Interpolation}. For interpolation of the global position corresponding to the odometry position at time $t_o$, four reference points on the global trajectory (blue points) are necessary. The interpolated global position $p_{g,t_0,inter}$ is on the resulting spline (green).}
    \label{fig:keyframeinterpolation}
\end{figure}

Our \textit{Keyframe Interpolation} is based on the use of the multidimensional spline curves implemented in \textit{Eigen}\footnote{https://eigen.tuxfamily.org}. The shape of the B-spline, defined by a set of GNSS keyframes, follows the equation:
\begin{equation}
    C(u) = \sum_{i=0}^{n}N_{i,p}(u)P_{i}
\end{equation}
A point $C(u)$ on the spline is described by the basis function $N_{i,p}(u)$ of degree $p$, the curve parameter $u$, and the reference point $P_{i}$. We implement a spline of degree $p = 3$, requiring four reference points. The reference points are selected by determining the two temporally closest GNSS frames before and after the current odometry point. A minimum spatial distance threshold between the GNSS frames used for interpolation ensures their well-spread distribution for reliable interpolation. The timestamp of the odometry keyframe is normalized to the range $[0,1]$, with the smallest GNSS timestamp serving as the minimum and the biggest GNSS timestamp as the maximum value. The interpolated position is calculated using the normalized value. \autoref{fig:keyframeinterpolation} illustrates the functionality of the \textit{Keyframe Interpolation} graphically.
\subsection{\textit{Rigid Alignment}}%
\label{subsec:rigidalignment}%
Given the interpolated trajectories, the odometry trajectory is optimally aligned to the global trajectory by applying a rigid transformation. To find the transformation, Umeyama's~\cite{umeyama} approach is used, minimizing the squared error between two point sets $\{\boldsymbol{x}_i\}$ and $\{\boldsymbol{y}_i\}$ with $n$ elements:
\begin{equation}
    e^2( \boldsymbol{R},\boldsymbol{t},s) = \frac{1}{n}\sum_{i=1}^{n} {|| \boldsymbol{y}_i - (s \boldsymbol{Rx}_i + \boldsymbol{t})||}^2
     \label{eq:umeyama_formula}
\end{equation}
The transformation consists of the rotation matrix $\boldsymbol R$, the translation vector $\boldsymbol t$, and a scaling factor $s$. However, as both trajectories are the same scale, only the rotational and translational parts of the transformation are applied to the odometry trajectory and the PCM.
\subsection{\textit{Rubber-Sheet Transformation}}%
\begin{figure}[!htb]
    \centering{% Define TUM colors
\definecolor{TUMBlue}{HTML}{0065BD}
\definecolor{TUMGray}{HTML}{808080}
\definecolor{TUMAccentOrange}{HTML}{E37222}
\definecolor{TUMAccentGreen}{HTML}{A2AD00}
\definecolor{TUMGreen}{rgb}{0.64, 0.68, 0.0}

\begin{tikzpicture}
    \begin{axis}[
        view={45}{30}, % 3D viewing angle
        colormap/jet, % Colormap for deviation data
        colorbar right, % Display colorbar on the right
        colorbar style={
            ylabel={Translation [\si{\meter}]},
            ytick={2.0, 3.0, 4.0, 5.0}, % Colorbar ticks
            width=0.15cm, % Thickness of the colorbar
        },
        xlabel={$x~[\si{\meter}]$},
        ylabel={$y~[\si{\meter}]$},
        zlabel={$z~[\si{\meter}]$},
        grid=both,
        width=0.85\linewidth,
        axis equal image,
        xmin=35, xmax=145,
        ymin=-510, ymax=-390,
        zmin=20, zmax=60,
    ]
    
    % Plot the original point cloud (gray)
    \addplot3[only marks, mark=*, mark size=0.5pt, color=TUMGray, opacity=0.7] 
        table [x index=0, y index=1, z index=2] {./fig/data/yas_hotel_align_cont.txt};
    
    % Plot the distorted point cloud with color mapping
    \addplot3[scatter, only marks, mark=*, mark size=0.5pt, scatter src=explicit, opacity=0.7] 
        table [x index=0, y index=1, z index=2, meta index=3] {./fig/data/yas_hotel_cont.txt};

    % Add a yellow dot at the specified coordinates
    \addplot3[only marks, mark=*, mark size=3pt, color=TUMAccentOrange] coordinates {(96.3519, -442.478, 33.9137)};

    % Add a plane (y = -410)
    %\addplot3[surf, samples=10, domain=-500:500, y domain=-200:200, shader=flat, opacity=0.3, %color=TUMAccentOrange]
    %    ({x}, -410, {z});

    \end{axis}
\end{tikzpicture}}
	\caption{Excerpt of a final PCM with color-coded point deformations by the \textit{Rubber-Sheet Transformation} at the YMC. Gray points represent the PCM after \textit{Rigid Alignment} with the orange mark being a selected CP subdividing the excerpt into tetrahedra.}
	\label{fig:yas_pcd_cont}
\end{figure}
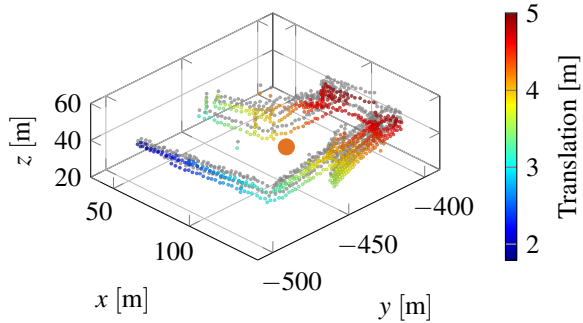

\label{subsec:rubbersheettrafo}%
The final step is a piecewise linear \textit{Rubber-Sheet Transformation}, derived from the research area of cartography~\cite{gillman1985triangulations}. Given a global reference map, it allows to correct distortions within a map by dividing it into subdivisions with different transformations~\cite{griffin}. As existing approaches are designed for standard 2D maps, we elevate the approach of Leitenstern et al.~\cite{leitenstern_flexmap_2024} to 3D and improve the CP selection and the triangulation method. \\%
The input parameters for the \textit{Rubber-Sheet Transformation} are the interpolated, global trajectory and the rigidly aligned odometry trajectory. CPs are automatically selected in the first step, leveraging the correspondence between both trajectories after the \textit{Keyframe Interpolation}. A configurable total amount of CPs is evenly distributed over the trajectories. However, if the GNSS standard deviation of a potential CP exceeds a threshold, the CP is skipped. The set of CPs is then extended by the eight points, forming a cuboid that encloses the trajectories with a configurable offset. In the next step, a Delaunay triangulation is created from the CPs, using the implementation of \textit{CGAL}~\cite{cgal_triag}, guaranteeing its uniqueness. The Delaunay triangulation is a well-defined and angle-optimal triangulation on a finite set of points, and thus, ideal for the given application~\cite{gillman1985triangulations}.
The transformation matrices of the resulting tetrahedra are calculated based on the CPs defining their corners:
\begin{equation}
    \boldsymbol{p_{g,i}} = \boldsymbol{T_j}\,\boldsymbol{p_{o,i}} \quad \quad \quad i \in \{k,\,l,\,m,\,q\}
    \label{eq:rs_vertices}
\end{equation}
Here, $T_j$ presents the linear transformation matrix of a single tetrahedron, defined by the four CPs $p_{g,i}$ and $p_{o,i}$ on the global and odometry trajectory, respectively. Leveraging this relationship for every vertex $i$ leads to a linear system of equations with \SI{16}{} variables, defining the components of $T_j$. To solve this system and thus compute the transformation matrices $T_j$ for the single tetrahedra, we employ the Householder QR  decomposition with full pivoting provided by \textit{Eigen}.
Finally, the odometry trajectory and the PCM are transformed by locating the corresponding tetrahedron for each point $\boldsymbol{x}$, and applying the corresponding transformation matrix:
\begin{equation}
    \boldsymbol{x}' = \boldsymbol{T_j}\,\boldsymbol{x}
    \label{eq:rubbersheet}
\end{equation}
While the transformation within a tetrahedron remains linear, the boundaries are adjusted in a way that preserves the topological properties of the map. \autoref{fig:yas_pcd_cont} exemplarily illustrates a resulting continuous PCM.
The whole pipeline is implemented as a standalone \textit{ROS 2}\footnote{https://ros.org/} package, leveraging \textit{RVIZ} to provide an interactive visualization of the final result.

%
% ======================= %
% === Results === %
% ======================= %
\section{\uppercase{Results}}
\label{sec:results}
\begin{figure*}
    % ===================
    % CP
    % ===================
    \begin{subfigure}[t]{0.49\linewidth}%
        \centering{\definecolor{TUMBlue}{HTML}{0065BD}
\definecolor{TUMSecondaryBlue}{HTML}{005293}
\definecolor{TUMSecondaryBlue2}{HTML}{003359}
\definecolor{TUMBlack}{HTML}{000000}
\definecolor{TUMWhite}{HTML}{FFFFFF}
\definecolor{TUMDarkGray}{HTML}{333333}
\definecolor{TUMGray}{HTML}{808080}
\definecolor{TUMLightGray}{HTML}{CCCCC6}
\definecolor{TUMAccentGray}{HTML}{DAD7CB}
\definecolor{TUMAccentOrange}{HTML}{E37222}
\definecolor{TUMAccentGreen}{HTML}{A2AD00}
\definecolor{TUMAccentLightBlue}{HTML}{98C6EA}
\definecolor{TUMAccentBlue}{HTML}{64A0C8}
\begin{tikzpicture}
    \begin{axis}[
        view={60}{30},
        xlabel={$x~[\si{\meter}]$},
        ylabel={$y~[\si{\meter}]$},
        zlabel={$z~[\si{\meter}]$},
        grid=both,
        width=\linewidth,
        axis equal image,
        xmin=-500,
        xmax=500,
        ymin=-900,
        ymax=900,
        zmin=-200,
        zmax=200
    ]

        % Umeyama aligned trajectory
        \addplot3[thick, color=TUMBlue] table [x index=0, y index=1, z index=2] {./fig/data/yas_combined_al.txt};
        % \addlegendentry{Aligned trajectory}

        % Triangular connections
        \input{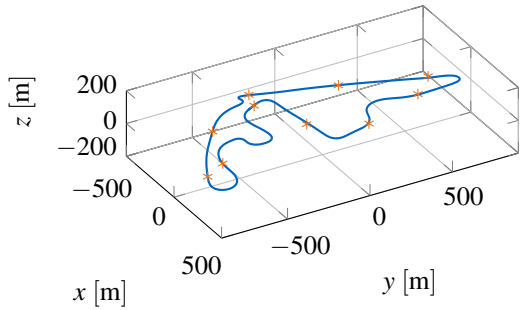}

    \end{axis}
\end{tikzpicture}}
	    \subcaption{Automatically selected CP (orange dots) on odometry trajectory.}
        \label{fig:yas_cp}
    \end{subfigure}%
    % ===================
    % Triangulation
    % ===================
    \begin{subfigure}[t]{0.49\linewidth}%
        \centering{\definecolor{TUMBlue}{HTML}{0065BD}
\definecolor{TUMSecondaryBlue}{HTML}{005293}
\definecolor{TUMSecondaryBlue2}{HTML}{003359}
\definecolor{TUMBlack}{HTML}{000000}
\definecolor{TUMWhite}{HTML}{FFFFFF}
\definecolor{TUMDarkGray}{HTML}{333333}
\definecolor{TUMGray}{HTML}{808080}
\definecolor{TUMLightGray}{HTML}{CCCCC6}
\definecolor{TUMAccentGray}{HTML}{DAD7CB}
\definecolor{TUMAccentOrange}{HTML}{E37222}
\definecolor{TUMAccentGreen}{HTML}{A2AD00}
\definecolor{TUMAccentLightBlue}{HTML}{98C6EA}
\definecolor{TUMAccentBlue}{HTML}{64A0C8}
\begin{tikzpicture}
    \begin{axis}[
        view={60}{30},
        xlabel={$x~[\si{\meter}]$},
        ylabel={$y~[\si{\meter}]$},
        zlabel={$z~[\si{\meter}]$},
        grid=both,
        width=\linewidth,
        axis equal image,
        xmin=-700,
        xmax=700,
        ymin=-1200,
        ymax=1200,
        zmin=-500,
        zmax=500,
    ]

        % Umeyama aligned trajectory
        \addplot3[thick, color=TUMBlue] table [x index=0, y index=1, z index=2] {./fig/data/yas_combined_al.txt};
        % \addlegendentry{Aligned trajectory}

        % Triangular connections
        \input{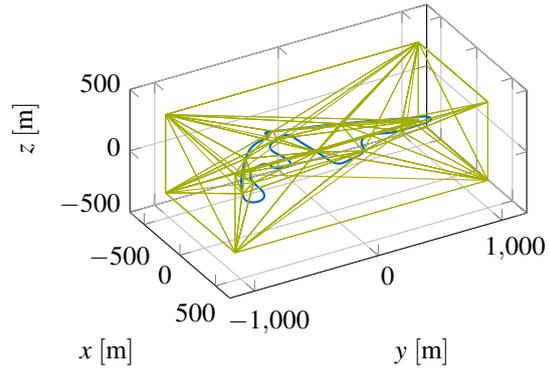}

    \end{axis}
\end{tikzpicture}}
	    \subcaption{Tetrahedra of resulting Delaunay triangulation.}
        \label{fig:yas_tet}
    \end{subfigure}
    \caption{Exemplary triangulation of the odometry trajectory (blue) using $n_{cp} = 10$ CP on the YMC. The size of the enclosing cuboid equals \SI{0.1}{} and \SI{10}{} times the expansion of the trajectory in $x/y$ and $z$, respectively.}
    \label{fig:yas_triag}
\end{figure*}

\begin{figure}[!htb]%
    \centering{\definecolor{TUMBlue}{HTML}{0065BD}
\definecolor{TUMSecondaryBlue}{HTML}{005293}
\definecolor{TUMSecondaryBlue2}{HTML}{003359}
\definecolor{TUMBlack}{HTML}{000000}
\definecolor{TUMWhite}{HTML}{FFFFFF}
\definecolor{TUMDarkGray}{HTML}{333333}
\definecolor{TUMGray}{HTML}{808080}
\definecolor{TUMLightGray}{HTML}{CCCCC6}
\definecolor{TUMAccentGray}{HTML}{DAD7CB}
\definecolor{TUMAccentOrange}{HTML}{E37222}
\definecolor{TUMAccentGreen}{HTML}{A2AD00}
\definecolor{TUMAccentLightBlue}{HTML}{98C6EA}
\definecolor{TUMAccentBlue}{HTML}{64A0C8}
\begin{tikzpicture}
    \begin{axis}[
        view={60}{30}, % 3D viewing angle
        colormap/jet, % Color map for surface
        colorbar right, % Colorbar on the right side
        colorbar style={
            ylabel={Deviation from GNSS [\si{\meter}]},
            ytick={0, 1, 2, 3},
            width=0.15cm, % sets the thickness of the colorbar
        },
        xlabel={$x~[\si{\meter}]$},
        ylabel={$y~[\si{\meter}]$},
        zlabel={$z~[\si{\meter}]$},
        grid=both,
        width=0.9\linewidth,
        axis equal image,
        xmin=-500,
        xmax=500,
        ymin=-750,
        ymax=750,
        zmin=-200,
        zmax=200
    ]
    % % Example plot using a 3D surface
    \addplot3[color=TUMGray, forget plot] 
        table [x index=0, y index=1, z index=2] {./fig/data/yas_source.txt};
    \addplot3[mesh, scatter, scatter src=explicit, mark=none] table [x index=0, y index=1, z index=2, meta index=3] {./fig/data/yas_combined_rs.txt};

    \end{axis}
\end{tikzpicture}}
	\caption{Final deviation of the odometry trajectory after \textit{Rubber-Sheet Transformation} using $n_{cp} = 200$ CP at the YMC.}
	\label{fig:yas_rs}
\end{figure}
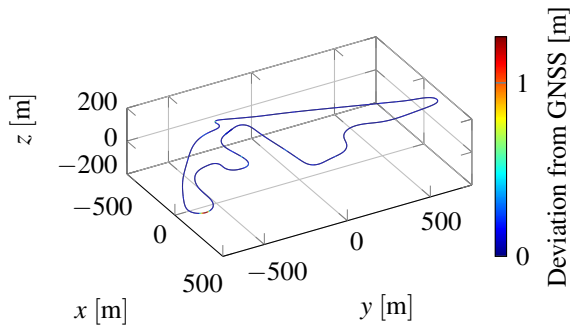

\begin{figure}[!htb]%
    \centering{\definecolor{TUMBlue}{HTML}{0065BD}
\definecolor{TUMSecondaryBlue}{HTML}{005293}
\definecolor{TUMSecondaryBlue2}{HTML}{003359}
\definecolor{TUMBlack}{HTML}{000000}
\definecolor{TUMWhite}{HTML}{FFFFFF}
\definecolor{TUMDarkGray}{HTML}{333333}
\definecolor{TUMGray}{HTML}{808080}
\definecolor{TUMLightGray}{HTML}{CCCCC6}
\definecolor{TUMAccentGray}{HTML}{DAD7CB}
\definecolor{TUMAccentOrange}{HTML}{E37222}
\definecolor{TUMAccentGreen}{HTML}{A2AD00}
\definecolor{TUMAccentLightBlue}{HTML}{98C6EA}
\definecolor{TUMAccentBlue}{HTML}{64A0C8}
\begin{tikzpicture}
    \begin{axis}[
        view={60}{30}, % 3D viewing angle
        colormap/jet, % Color map for surface
        colorbar right, % Colorbar on the right side
        colorbar style={
            ylabel={Deviation from GNSS [\si{\meter}]},
            ytick={0, 1, 2, 3},
            width=0.15cm, % sets the thickness of the colorbar
        },
        xlabel={$x~[\si{\meter}]$},
        ylabel={$y~[\si{\meter}]$},
        zlabel={$z~[\si{\meter}]$},
        grid=both,
        width=0.6\linewidth,
        scale only axis,
        axis equal image,
        xmin=-100,
        xmax=400,
        ymin=-100,
        ymax=450,
        zmin=-100,
        zmax=100
    ]
    % % Example plot using a 3D surface
    \addplot3[color=TUMGray, forget plot] 
        table [x index=0, y index=1, z index=2] {./fig/data/kitti_source.txt};
    \addplot3[mesh, scatter, scatter src=explicit, mark=none] table [x index=0, y index=1, z index=2, meta index=3] {./fig/data/kitti_combined_rs.txt};

    \end{axis}
\end{tikzpicture}}
	\caption{Final deviation of the odometry trajectory after \textit{Rubber-Sheet Transformation} using $n_{cp} = 150$ CP on sequence 00 of the KITTI dataset~\cite{geiger_are_2012}.}
	\label{fig:kitti_rs}
\end{figure}
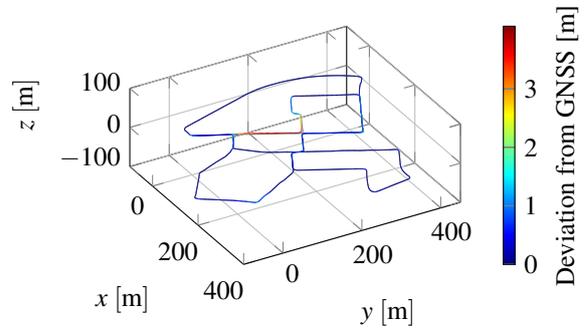

Both datasets, the YMC, and the KITTI dataset, comprise RTK-corrected GNSS positions and LiDAR point clouds, used as input to \textit{FlexCloud}. As georeferencing requires the GNSS positions in a global frame, raw data of the KITTI dataset is used. However, it is noted at this point that due to the urban environment in sequence 00 of the KITTI dataset, the GNSS positions feature high standard deviations for several sections. The GNSS positions recorded at the YMC are post-processed and show standard deviation in the centimeter range, except for short occlusions. The initial odometry trajectory and corresponding PCM are created as described in \autoref{sec:meth}. The evaluation consists of two parts: First, we present quantitative results on the deviation of the final odometry to the GNSS trajectory. For this, the Euclidean norm is computed for each point on the odometry trajectory and its corresponding point on the GNSS trajectory after the \textit{Keyframe Interpolation}. In contrast to the \textit{Rubber-Sheet} transformation, which only uses selected CPs, the evaluation is conducted on all trajectory points without excluding points with high standard deviation. The second part of the evaluation is conducted on publicly available satellite imagery. This enables a qualitative evaluation with an external data source. \\%
The resulting trajectories after the \textit{Keyframe Interpolation} consist of \SI{2560}{} and \SI{1550}{} poses for the YMC and the KITTI dataset, respectively. To improve the understanding of the pipeline, intermediate results are presented for the YMC.
\begin{figure}[!htb]
    \centering{\definecolor{TUMBlue}{HTML}{0065BD}
\definecolor{TUMSecondaryBlue}{HTML}{005293}
\definecolor{TUMSecondaryBlue2}{HTML}{003359}
\definecolor{TUMBlack}{HTML}{000000}
\definecolor{TUMWhite}{HTML}{FFFFFF}
\definecolor{TUMDarkGray}{HTML}{333333}
\definecolor{TUMGray}{HTML}{808080}
\definecolor{TUMLightGray}{HTML}{CCCCC6}
\definecolor{TUMAccentGray}{HTML}{DAD7CB}
\definecolor{TUMAccentOrange}{HTML}{E37222}
\definecolor{TUMAccentGreen}{HTML}{A2AD00}
\definecolor{TUMAccentLightBlue}{HTML}{98C6EA}
\definecolor{TUMAccentBlue}{HTML}{64A0C8}
\begin{tikzpicture}
    \begin{axis}[
        view={60}{30}, % 3D viewing angle
        colormap/jet, % Color map for surface
        colorbar right, % Colorbar on the right side
        colorbar style={
            ylabel={Deviation from GNSS [\si{\meter}]},
            ytick={0, 5, 10, 15},
            width=0.15cm, % sets the thickness of the colorbar
        },
        xlabel={$x~[\si{\meter}]$},
        ylabel={$y~[\si{\meter}]$},
        zlabel={$z~[\si{\meter}]$},
        grid=both,
        width=0.9\linewidth,
        axis equal image,
        xmin=-500,
        xmax=500,
        ymin=-750,
        ymax=750,
        zmin=-200,
        zmax=200
    ]
    % % Example plot using a 3D surface
    \addplot3[color=TUMGray, forget plot] 
        table [x index=0, y index=1, z index=2] {./fig/data/yas_source.txt};
    \addplot3[mesh, scatter, scatter src=explicit, mark=none] table [x index=0, y index=1, z index=2, meta index=3] {./fig/data/yas_combined_al.txt};

    \end{axis}
\end{tikzpicture}}
	\caption{Deviation of the odometry to the GNSS trajectory (gray) after \textit{Rigid Alignment} at the YMC.}
	\label{fig:yas_umeyama}
\end{figure}
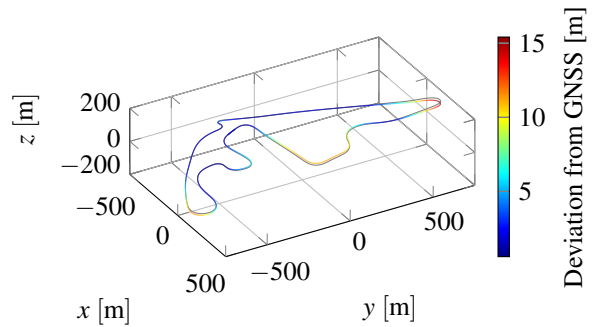
\autoref{fig:yas_umeyama} shows the deviation of the odometry to the GNSS trajectory after the \textit{Rigid Alignment} step.
Although the two trajectories are optimally aligned with a rigid transformation at this point, the deviation exceeds \SI{15}{\meter} in certain sections. However, as the deviation is not constant, this highlights the distortion of the trajectory and thus the need for the \textit{Rubber-Sheet Transformation}. \\%
\autoref{fig:yas_triag} illustrates the triangulation of the odometry trajectory in the \textit{Rubber-Sheet Transformation}.
First, the CPs are selected based on the configuration, and the maximum allowed standard deviation of the GNSS positions (\autoref{fig:yas_cp}). In the second step, the corner points of the enclosing cuboid are added, and the Delaunay triangulation is computed. \\%
Leveraging the transformation matrices computed from the triangulation, the \textit{Rubber-Sheet Transformation} is applied according to \autoref{eq:rubbersheet} to the odometry trajectory and the PCM. The deviation of the final odometry to the GNSS trajectory using \SI{200}{} CPs for the YMC is illustrated in \autoref{fig:yas_rs}. 
Comparing the final deviation to the intermediate results after the \textit{Rigid Alignment} step (\autoref{fig:yas_umeyama}), the benefit of the \textit{Rubber-Sheet Transformation} in rectifying the trajectory is clearly visible. Except for a short section ($x \approx \SI{400}{\meter}, y \approx \SI{-800}{\meter}$), where no CPs are selected following high GNSS standard deviations, the deviation falls below \SI{1}{\meter} for the entire trajectory. The results for sequence 00 of the KITTI dataset in \autoref{fig:kitti_rs} show similar results.
Analogous to the YMC, the deviation increases within a section in the middle of the trajectory due to high GNSS standard deviations preventing the CP selection. As this shows the influence of the amount and location of the CPs, \autoref{tab:quant} compares the results of different quantities of CPs statistically.
\begin{table}[!htb]%
    \caption{Quantitative results for the mean absolute error (MAE), the standard deviation (STDDEV), and the maximum value (MAX) of the deviation of the georeferenced odometry trajectory to the GNSS trajectory. Results are provided for different amounts of CP $n_{cp}$ with a standard deviation threshold of \SI{0.05}{\meter} and \SI{0.25}{\meter} for the YMC and the KITTI dataset, respectively.}
    \centering%
    \begin{tabular}{|c|c|c|c|c|c|}%
        \hline%
        \multicolumn{2}{|c|}{\multirow{2}{0.25\linewidth}{\textbf{Dataset\,$\mid\,n_{cp}$}}} & \multicolumn{3}{|c|}{\textbf{Metric in \si{\meter}}} \\%
        \cline{3-5}
        \multicolumn{2}{|c|}{} & \textbf{MAE} & \textbf{STDDEV} & \textbf{MAX} \\%
        \hline%
        \multirow{4}{0.25\linewidth}{Yas Marina \newline Circuit} & \SI{10}{} & \SI{1.71}{} & \SI{1.59}{} & \SI{9.11}{} \\%
                                               & \SI{50}{} & \SI{0.25}{} & \SI{0.32}{} & \SI{2.56}{} \\%
                                               & \SI{100}{} & \SI{0.13}{} & \SI{0.16}{} & \SI{1.43}{} \\%
                                               & \SI{200}{} & \SI{0.08}{} & \SI{0.11}{} & \SI{1.27}{} \\%
        \hline%
        \multirow{4}{0.25\linewidth}{KITTI 00} & \SI{10}{} & \SI{1.26}{} & \SI{0.77}{} & \SI{4.03}{} \\%
                                            & \SI{50}{} & \SI{0.60}{} & \SI{0.77}{} & \SI{3.71}{} \\%
                                            & \SI{150}{} & \SI{0.47}{} & \SI{0.83}{} & \SI{4.07}{} \\%
        \hline
    \end{tabular}%
    \label{tab:quant}
\end{table}
For both datasets, the MAE decreases with an increasing amount of CPs, leading to an average error of \SI{0.08}{\meter} and \SI{0.47}{\meter} for the YMC and the KITTI dataset, respectively. This is expected as the distance between successive CPs decreases. The larger value for the KITTI dataset follows from an overall worse quality of the GNSS positions. While STDDEV and MAX also decrease with an increasing amount of CPs for the YMC, this is not observable for the KITTI dataset. This is due to the long section with insufficient GNSS accuracy in the middle of the trajectory. It is noticeable that for $n_{cp} = 10$, the KITTI results are superior to the YMC despite the worse quality of the GNSS trajectory. This can be explained by the shorter overall length of the KITTI dataset, causing a reduced distance between consecutive CPs. \\%
We leverage aerial photography to evaluate the results independently of the GNSS positions, which are also used as input to \textit{FlexCloud}. Although this does not enable a quantitative but a qualitative evaluation, it allows further investigation of the performance of \textit{FlexCloud}, especially within the sections with high GNSS uncertainty. For the YMC, the south part is visualized over satellite imagery from ESRI~\cite{esri_satellite}~(\autoref{fig:yas_south_esri}).
\begin{figure}[!htb]
  	\centering
   	\includegraphics[width=\linewidth]{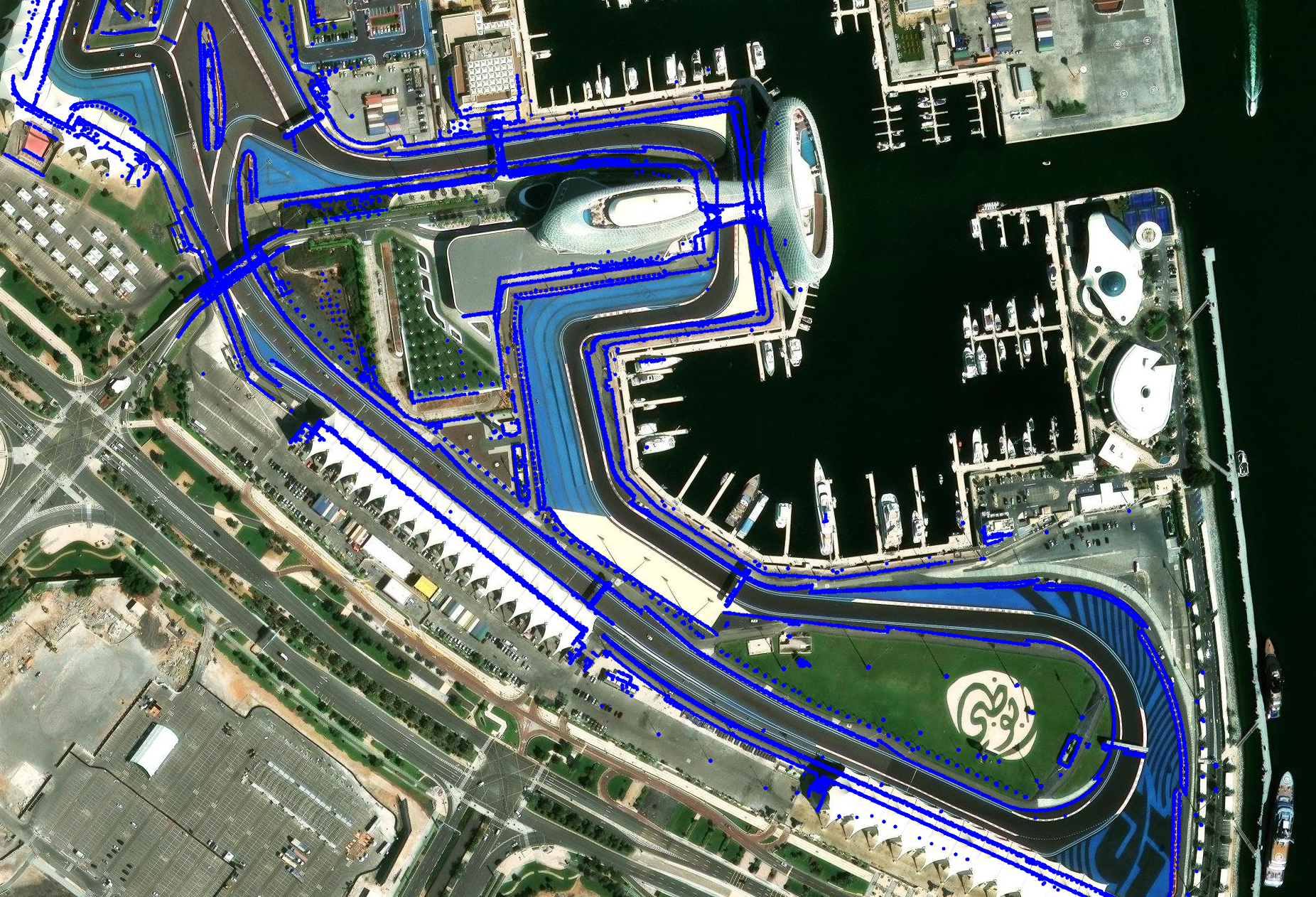} \\%
   	\caption{Visualization of PCM of the YMC on satellite imagery from ESRI~\cite{esri_satellite}}%
   	\label{fig:yas_south_esri}%
\end{figure}
A visual comparison indicates a strong overlap between the PCM and the satellite image, especially in the turn in the southeast of the image. Although \autoref{fig:yas_rs} shows an increased deviation between the trajectories in this section, no deterioration in the overlap is visible. This proves the ability of \textit{FlexCloud} to enable reliable georeferencing even for short sections with insufficient GNSS coverage. \\%
As the KITTI dataset was recorded in the state of Baden-Württemberg, Germany, public orthophotos with a resolution of \SI{20}{\centi \meter} are available for evaluation (\autoref{fig:kitti_ortho}). 
\begin{figure}[!htb]
  	\centering
   	\includegraphics[width=\linewidth]{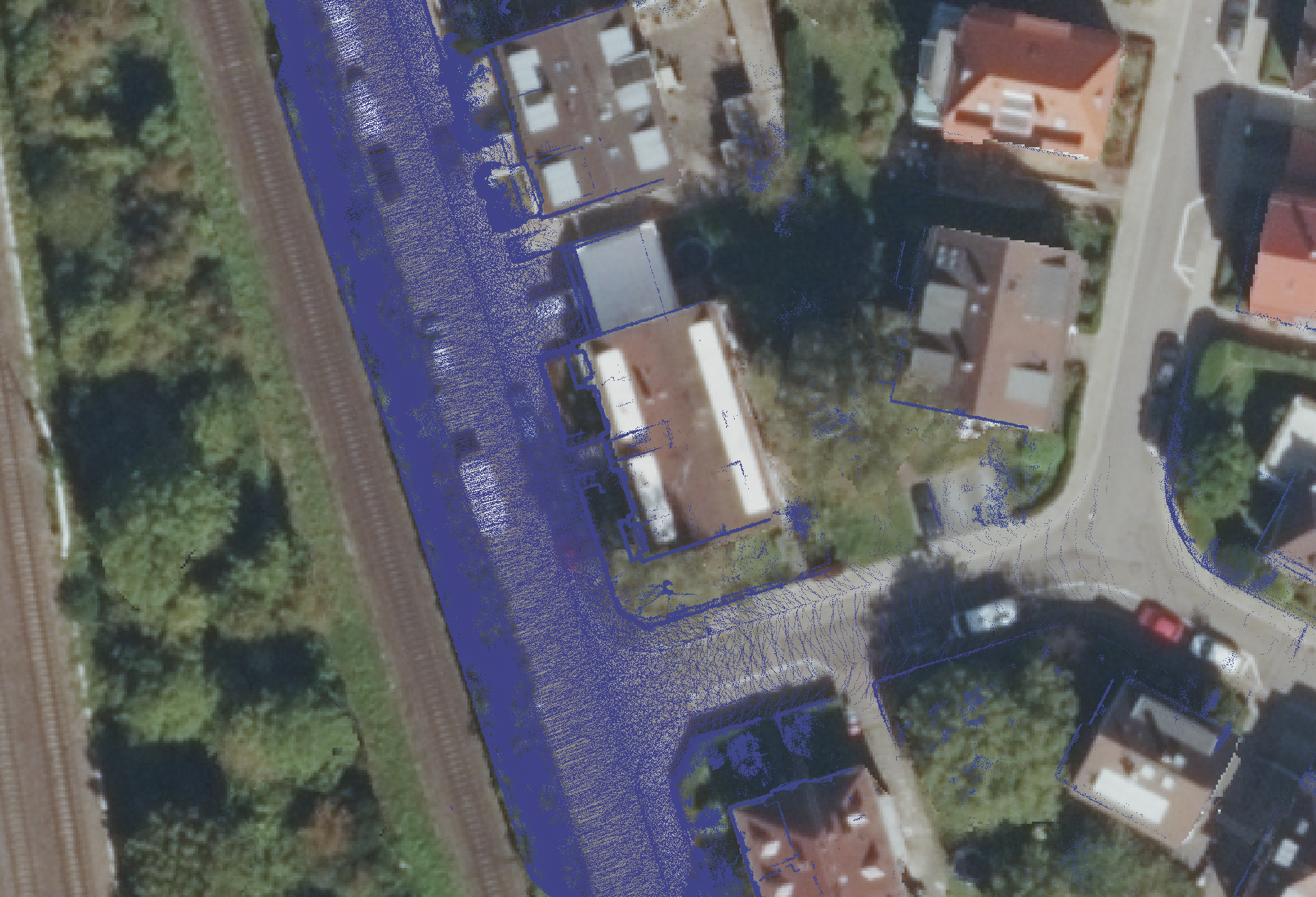} \\%
   	\caption{Visualization of PCM of KITTI sequence 00 over orthophotos of the State Office for
Geoinformation and Rural Development (LGL)\protect\footnotemark of Baden-Württemberg.}%
   	\label{fig:kitti_ortho}%
\end{figure}
\footnotetext{https://www.lgl-bw.de/}
The shown excerpt is taken from the southwest of the trajectory ($x \approx \SI{-60}{\meter}, y \approx \SI{70}{\meter}$) featuring good GNSS coverage. From the high resolution of the orthophoto, it is visible that the road boundaries and building structures of the PCM accurately match the orthophoto. This indicates that PCMs georeferenced with \textit{FlexCloud} can reach the required accuracy for HD maps given a global trajectory with sufficient accuracy. 
%
% ======================= %
% === Discussion === %
% ======================= %
\section{\uppercase{Discussion}}
\label{sec:discussion}
The presented results demonstrate that \textit{FlexCloud} can accurately georeference a given local PCM based on the corresponding odometry and GNSS trajectory. Nevertheless, our approach still features limitations and areas for improvement, which are discussed in this section. By introducing the \textit{Keyframe Interpolation}, we can directly utilize the GNSS trajectory recorded by the MMS to generate CPs for georeferencing. Although this avoids the need for additional measurements, it has the following limitations: To reach an accuracy sufficient for autonomous driving applications, the MMS needs to be equipped with an expensive RTK-corrected GNSS receiver. Furthermore, the \textit{Keyframe Interpolation} requires proper time synchronization between the GNSS receiver and the LiDAR, as the interpolation leverages the timestamp differences between the trajectory positions. \\%
The \textit{Rubber-Sheet Transformation} leverages several CPs, a subset of all positions of the trajectories, to compute the transformation of the PCM. By excluding GNSS positions with high uncertainty from being selected as CP, these sections can still be georeferenced to a certain extent. However, this requires a reliable standard deviation estimation of the GNSS receiver. Furthermore, our current implementation does not utilize the actual values of the GNSS standard deviation but only excludes high values based on a fixed threshold. Hence, additional focus must be spent on post-processing the GNSS positions, e.g., using curvature correction.
Another point to be investigated is the shape structure surrounding the trajectories/map during the \textit{Rubber-Sheet Transformation}. Although the rectangular cuboid used within this work shows robust results, limitations arise, e.g., for trajectories/maps, where the surrounding cuboid features large variations in the side lengths. An alternative approach would be the replacement of the cuboid with a polygon-like shape.
\section{\uppercase{Conclusion}}
\label{sec:conclusion}
\textit{FlexCloud} provides a framework for direct georeferencing and drift-correction of local PCMs created from SLAM. Following the modular design, our approach can be combined with different SLAM pipelines, utilizing only the generated odometry trajectory and local PCM. By implementing the \textit{Keyframe Interpolation}, the pipeline can directly leverage the GNSS positions recorded with an MMS, avoiding the need for additional measurements. The \textit{Rubber-Sheet Transformation} enables rectification of the local PCM and georeferencing of short sections without accurate GNSS positions. Although our approach still features limitations, it provides the opportunity to georeference PCMs with high accuracy, allowing their usage for map-based localization algorithms. Future work on the pipeline needs to focus on the boundary shape used for the \textit{Rubber-Sheet Transformation} and improved processing of the initial GNSS positions based on their standard deviation.
\section*{\uppercase{Acknowledgements}}
As the first author, Maximilian Leitenstern initiated and designed the paper's structure. He contributed to designing and implementing the overall pipeline concept. Christian Bolea-Schaser and Marko Alten contributed to designing and implementing the Rubber-Sheet Transformation and the Keyframe Interpolation during their student research project, respectively. Dominik Kulmer, Marcel Weinmann, and Markus Lienkamp revised the paper critically for important intellectual content. Markus Lienkamp gives final approval for the version to be published and agrees to all aspects of the work. As a guarantor, he accepts responsibility for the overall integrity of the paper.
\bibliographystyle{apalike}
{\small
\bibliography{references}}

\begin{thebibliography}{}

\bibitem[Bao et~al., 2022]{bao_high-definition_2022}
Bao, Z., Hossain, S., Lang, H., and Lin, X. (2022).
\newblock High-{Definition} {Map} {Generation} {Technologies} {For}
  {Autonomous} {Driving}.
\newblock {\em ArXiv}, abs/2206.05400.

\bibitem[Bao et~al., 2023]{bao_review_2023}
Bao, Z., Hossain, S., Lang, H., and Lin, X. (2023).
\newblock A review of high-definition map creation methods for autonomous
  driving.
\newblock {\em Engineering Applications of Artificial Intelligence},
  122:106125.

\bibitem[Chen et~al., 2019]{chen_suma_2019}
Chen, X., Milioto, A., Palazzolo, E., Giguere, P., Behley, J., and Stachniss,
  C. (2019).
\newblock {SuMa}++: {Efficient} {LiDAR}-based {Semantic} {SLAM}.
\newblock In {\em 2019 {IEEE}/{RSJ} {International} {Conference} on
  {Intelligent} {Robots} and {Systems} ({IROS})}. IEEE.

\bibitem[Esri, 2024]{esri_satellite}
Esri (2024).
\newblock Arcgis world imagery.
\newblock Available at:
  \url{https://www.arcgis.com/apps/mapviewer/index.html?layers=10df2279f9684e4a9f6a7f08febac2a9}
  (accessed on 2024-11-29).

\bibitem[Geiger et~al., 2013]{geiger_vision_2013}
Geiger, A., Lenz, P., Stiller, C., and Urtasun, R. (2013).
\newblock Vision meets robotics: {The} {KITTI} dataset.
\newblock {\em The International Journal of Robotics Research},
  32(11):1231--1237.
\newblock \_eprint: https://doi.org/10.1177/0278364913491297.

\bibitem[Geiger et~al., 2012]{geiger_are_2012}
Geiger, A., Lenz, P., and Urtasun, R. (2012).
\newblock Are we ready for autonomous driving? {The} {KITTI} vision benchmark
  suite.
\newblock In {\em 2012 {IEEE} {Conference} on {Computer} {Vision} and {Pattern}
  {Recognition}}, pages 3354--3361.

\bibitem[Gillman, 1985]{gillman1985triangulations}
Gillman, D. (1985).
\newblock Triangulations for rubber-sheeting.
\newblock In {\em Proceedings of 7th International symposium on computer
  assisted cartography (AutoCarto 7)}, volume 199.

\bibitem[Groves, 2013]{groves_principles_2013}
Groves, P. (2013).
\newblock {\em Principles of {GNSS}, {Inertial}, and {Multisensor} {Integrated}
  {Navigation} {Systems}, {Second} {Edition}}.
\newblock Artech.

\bibitem[Hariz et~al., 2021]{hariz_direct_2021}
Hariz, F., Souifi, H., Leblanc, R., Bouslimani, Y., Ghribi, M., Langin, E., and
  Mccarthy, D. (2021).
\newblock Direct {Georeferencing} {3D} {Points} {Cloud} {Map} {Based} on {SLAM}
  and {Robot} {Operating} {System}.
\newblock In {\em 2021 {IEEE} {International} {Symposium} on {Robotic} and
  {Sensors} {Environments} ({ROSE})}, pages 1--6.

\bibitem[Hess et~al., 2016]{hess_2016}
Hess, W., Kohler, D., Rapp, H., and Andor, D. (2016).
\newblock Real-time loop closure in 2d lidar slam.
\newblock In {\em 2016 IEEE International Conference on Robotics and Automation
  (ICRA)}, pages 1271--1278.

\bibitem[Jamin et~al., 2024]{cgal_triag}
Jamin, C., Pion, S., and Teillaud, M. (2024).
\newblock {3D} triangulations.
\newblock In {\em {CGAL} User and Reference Manual}. {CGAL Editorial Board},
  {6.0.1} edition.

\bibitem[Jeong et~al., 2022]{jeong_tutorial_2022}
Jeong, J., Yoon, J.~Y., Lee, H., Darweesh, H., and Sung, W. (2022).
\newblock Tutorial on {High}-{Definition} {Map} {Generation} for {Automated}
  {Driving} in {Urban} {Environments}.
\newblock {\em Sensors}, 22(18).

\bibitem[Khoche et~al., 2022]{khoche_semantic_2022}
Khoche, A., Wozniak, M.~K., Duberg, D., and Jensfelt, P. (2022).
\newblock Semantic {3D} {Grid} {Maps} for {Autonomous} {Driving}.
\newblock {\em 2022 IEEE 25th International Conference on Intelligent
  Transportation Systems (ITSC)}, pages 2681--2688.

\bibitem[Koide et~al., 2019]{koide_portable_2019}
Koide, K., Miura, J., and Menegatti, E. (2019).
\newblock A portable three-dimensional {LIDAR}-based system for long-term and
  wide-area people behavior measurement.
\newblock {\em International Journal of Advanced Robotic Systems},
  16(2):1729881419841532.
\newblock \_eprint: https://doi.org/10.1177/1729881419841532.

\bibitem[Koide et~al., 2021]{koide_interactive_2021}
Koide, K., Miura, J., Yokozuka, M., Oishi, S., and Banno, A. (2021).
\newblock Interactive {3D} {Graph} {SLAM} for {Map} {Correction}.
\newblock {\em IEEE Robotics and Automation Letters}, 6(1):40--47.

\bibitem[{Lasse Klingbeil}, 2023]{lasse_klingbeil_georeferencing_2023}
{Lasse Klingbeil} (2023).
\newblock {\em Georeferencing of {Mobile} {Mapping} {Data}}.
\newblock {PhD} {Thesis}, Rheinische Friedrich-Wilhelms-Universität Bonn.

\bibitem[Leitenstern et~al., 2024]{leitenstern_flexmap_2024}
Leitenstern, M., Sauerbeck, F., Kulmer, D., and Betz, J. (2024).
\newblock {FlexMap} {Fusion}: {Georeferencing} and {Automated} {Conflation} of
  {HD} {Maps} with {OpenStreetMap}.
\newblock In {\em 35th {IEEE} {Intelligent} {Vehicles} {Symposium}, {IV} 2024},
  {IEEE} {Intelligent} {Vehicles} {Symposium}, {Proceedings}, pages 272--278.
  Institute of Electrical and Electronics Engineers Inc.

\bibitem[Liu et~al., 2021]{liu_georeferencing_2021}
Liu, W., Li, Z., Sun, S., Du, H., and Sotelo, M.~A. (2021).
\newblock Georeferencing kinematic modeling and error correction of terrestrial
  laser scanner for {3D} scene reconstruction.
\newblock {\em Automation in Construction}, 126:103673.

\bibitem[Marvin S.~White and Griffin, 1985]{griffin}
Marvin S.~White, J. and Griffin, P. (1985).
\newblock Piecewise linear rubber-sheet map transformation.
\newblock {\em The American Cartographer}, 12(2):123--131.

\bibitem[Oria-Aguilera et~al., 2018]{oria-aguilera_mobile_2018}
Oria-Aguilera, H., Alvarez-Perez, H., and Garcia-Garcia, D. (2018).
\newblock Mobile {LiDAR} {Scanner} for the {Generation} of {3D} {Georeferenced}
  {Point} {Clouds}.
\newblock In {\em 2018 {IEEE} {International} {Conference} on
  {Automation}/{XXIII} {Congress} of the {Chilean} {Association} of {Automatic}
  {Control} ({ICA}-{ACCA})}, pages 1--6.

\bibitem[Osada et~al., 2017]{osada_direct_2017}
Osada, E., Sośnica, K., Borkowski, A., Owczarek-Wesołowska, M., and Gromczak,
  A. (2017).
\newblock A {Direct} {Georeferencing} {Method} for {Terrestrial} {Laser}
  {Scanning} {Using} {GNSS} {Data} and the {Vertical} {Deflection} from
  {Global} {Earth} {Gravity} {Models}.
\newblock {\em Sensors}, 17(7).

\bibitem[Paffenholz, 2012]{paffenholz_direct_2012}
Paffenholz, J.-A. (2012).
\newblock {\em Direct geo-referencing of {3D} point clouds with {3D}
  positioning sensors}.
\newblock {PhD} {Thesis}, Leibniz-Universität Hannover.

\bibitem[Sauerbeck et~al., 2023]{sauerbeck2023}
Sauerbeck, F., Kulmer, D., Pielmeier, M., Leitenstern, M., Weiß, C., and Betz,
  J. (2023).
\newblock Multi-lidar localization and mapping pipeline for urban autonomous
  driving.
\newblock In {\em 2023 IEEE SENSORS}, pages 1--4.

\bibitem[Schuhmacher and Boehm, 2005]{schuhmacher_georeferencing_2005}
Schuhmacher, S. and Boehm, J. (2005).
\newblock Georeferencing of {Terrestrial} {Laserscanner} {Data} for
  {Applications} in {Architectural} {Modelling}.
\newblock {\em International Archives of Photogrammetry, Remote Sensing and
  Spatial Information Sciences}, 36.

\bibitem[Seif and Hu, 2016]{seif_autonomous_2016}
Seif, H.~G. and Hu, X. (2016).
\newblock Autonomous {Driving} in the {iCity}—{HD} {Maps} as a {Key}
  {Challenge} of the {Automotive} {Industry}.
\newblock {\em Engineering}, 2(2):159--162.

\bibitem[Shan and Englot, 2018]{shan_lego-loam_2018}
Shan, T. and Englot, B. (2018).
\newblock {LeGO}-{LOAM}: {Lightweight} and {Ground}-{Optimized} {Lidar}
  {Odometry} and {Mapping} on {Variable} {Terrain}.
\newblock In {\em 2018 {IEEE}/{RSJ} {International} {Conference} on
  {Intelligent} {Robots} and {Systems} ({IROS})}, pages 4758--4765.

\bibitem[Shan et~al., 2020]{shan_lio-sam_2020}
Shan, T., Englot, B., Meyers, D., Wang, W., Ratti, C., and Rus, D. (2020).
\newblock {LIO}-{SAM}: {Tightly}-coupled {Lidar} {Inertial} {Odometry} via
  {Smoothing} and {Mapping}.
\newblock \_eprint: 2007.00258.

\bibitem[Srinara et~al., 2023]{srinara_strategy_2023}
Srinara, S., Chiu, Y.-T., Chen, J.-A., Chiang, K.-W., Tsai, M.-L., and
  El-Sheimy, N. (2023).
\newblock Strategy on {High}-{Definition} {Point} {Cloud} {Map} {Creation} for
  {Autonomous} {Driving} in {Highway} {Environments}.
\newblock {\em The International Archives of the Photogrammetry, Remote Sensing
  and Spatial Information Sciences}, XLVIII-1/W2-2023:849--854.

\bibitem[Umeyama, 1991]{umeyama}
Umeyama, S. (1991).
\newblock Least-squares estimation of transformation parameters between two
  point patterns.
\newblock {\em IEEE Transactions on Pattern Analysis and Machine Intelligence},
  13(4):376--380.

\bibitem[Vizzo et~al., 2023]{vizzo_kiss-icp_2023}
Vizzo, I., Guadagnino, T., Mersch, B., Wiesmann, L., Behley, J., and Stachniss,
  C. (2023).
\newblock {KISS}-{ICP}: {In} {Defense} of {Point}-to-{Point} {ICP} {Simple},
  {Accurate}, and {Robust} {Registration} {If} {Done} the {Right} {Way}.
\newblock {\em IEEE Robotics and Automation Letters}, 8(2):1029--1036.
\newblock Publisher: Institute of Electrical and Electronics Engineers (IEEE).

\bibitem[Wilkinson et~al., 2015]{wilkinson_dual-antenna_2015}
Wilkinson, B., Mohamed, A., and Dewitt, B. (2015).
\newblock Dual-{Antenna} {Terrestrial} {Laser} {Scanner} {Georeferencing}
  {Using} {Auxiliary} {Photogrammetric} {Observations}.
\newblock {\em Remote Sensing}, 7(9):11621--11638.

\bibitem[Wilkinson et~al., 2010]{wilkinson_novel_2010}
Wilkinson, B.~E., Mohamed, A.~H., Dewitt, B.~A., and Seedahmed, G.~H. (2010).
\newblock A {Novel} {Approach} to {Terrestrial} {Lidar} {Georeferencing}.
\newblock {\em Photogrammetric Engineering \& Remote Sensing}, 76(6):683--690.

\bibitem[Xu et~al., 2021]{xu_fast-lio2_2021}
Xu, W., Cai, Y., He, D., Lin, J., and Zhang, F. (2021).
\newblock {FAST}-{LIO2}: {Fast} {Direct} {LiDAR}-inertial {Odometry}.
\newblock \_eprint: 2107.06829.

\bibitem[Xu and Zhang, 2021]{xu_fast-lio_2021}
Xu, W. and Zhang, F. (2021).
\newblock {FAST}-{LIO}: {A} {Fast}, {Robust} {LiDAR}-inertial {Odometry}
  {Package} by {Tightly}-{Coupled} {Iterated} {Kalman} {Filter}.
\newblock \_eprint: 2010.08196.

\bibitem[Zhang and Singh, 2014]{zhang_loam_2014}
Zhang, J. and Singh, S. (2014).
\newblock {LOAM} : {Lidar} {Odometry} and {Mapping} in real-time.
\newblock {\em Robotics: Science and Systems Conference (RSS)}, pages 109--111.

\end{thebibliography}
\end{document}